\documentclass[letterpaper, 10 pt, conference]{ieeeconf}  %

\IEEEoverridecommandlockouts                              %

\usepackage{graphics} %
\usepackage{epsfig} %
\usepackage{times} %
\usepackage{amsmath} %
\usepackage{amssymb}  %
\usepackage[ruled]{algorithm2e}

\usepackage{multirow}
\usepackage{multicol}
\usepackage{graphicx}
\usepackage{subcaption}
\usepackage{colortbl}
\usepackage{tabularx}
\usepackage{makecell}
\usepackage[table]{xcolor}
\usepackage{float}
\usepackage{hyperref}
\usepackage{cleveref}
\definecolor{velvet}{HTML}{E37222}
\hypersetup{
  colorlinks   = true, 
  urlcolor     = velvet, 
  linkcolor    = velvet, 
  citecolor   = velvet 
}
\newcommand{\customfootnotetext}[2]{{%
  \renewcommand{\thefootnote}{#1}%
  \footnotetext[0]{#2}}}%

\title{\LARGE \bf
Object Pose Estimation through Dexterous Touch
}

\author{Amir Hossein Shahidzadeh$^{1*}$, Jiyue Zhu$^{2*}$, Kezhou Chen$^{2}$, \\Sha Yi$^{2}$, Cornelia Ferm\"{u}ller$^{1}$, Yiannis Aloimonos$^{1}$ and Xiaolong Wang$^{2}$
<-this %
\thanks{}%
\thanks{$^{1}$Albert Author is with Faculty of Electrical Engineering, Mathematics and Computer Science,
        University of Twente, 7500 AE Enschede, The Netherlands
        {\tt\small albert.author@papercept.net}}%
\thanks{$^{2}$Bernard D. Researcheris with the Department of Electrical Engineering, Wright State University,
        Dayton, OH 45435, USA
        {\tt\small b.d.researcher@ieee.org}}%
}

\begin{document}

\twocolumn[{%
\renewcommand\twocolumn[1][]{#1}%

\maketitle
\thispagestyle{empty}
\pagestyle{empty}

}]

\customfootnotetext{*}{Indicates equal contribution.}
\customfootnotetext{1}{University of Maryland, College Park}
\customfootnotetext{2}{UC San Diego}

\begin{abstract}
Robust object pose estimation is essential for manipulation and interaction tasks in robotics, particularly in scenarios where visual data is limited or sensitive to lighting, occlusions, and appearances. Tactile sensors often offer limited and local contact information, making it challenging to reconstruct the pose from partial data. Our approach uses sensorimotor exploration to actively control a robot hand to interact with the object. We train with Reinforcement Learning (RL) to explore and collect tactile data. The collected 3D point clouds are used to iteratively refine the object’s shape and pose. In our setup, one hand holds the object steady while the other performs active exploration. We show that our method can actively explore an object’s surface to identify critical pose features without prior knowledge of the object's geometry. Supplementary material and more demonstrations will be provided at  \href{https://amirshahid.github.io/BimanualTactilePose}{https://amirshahid.github.io/BimanualTactilePose}.
\end{abstract}

\vspace{-0.5cm}
\section{Introduction}
\label{sec:introduction}

Accurate object pose estimation is essential for dexterous robotic manipulation, enabling reliable interactions across a range of environments~\cite{ilonen2014three, labbe2022megapose}. Traditional vision-based approaches have been effective but are limited by their sensitivity to lighting, occlusions, and material properties~\cite{bian2023transtouch, bauer2024challenges}. External cameras often introduce bulk and complexity when integrated with dexterous robotic hands. In contrast, tactile sensors, such as force-sensing resistors (FSRs), offer a compact and low-cost alternative suitable for embedding directly into robot hand fingertips~\cite{yin2023rotating}. However, tactile sensors~\cite{pai2023tactofind} provide only sparse, local data, giving only isolated contact points rather than dense spatial information. To achieve the level of geometric detail required for accurate pose estimation, tactile approaches need active, long-horizon exploration to build up a comprehensive object model through interaction.

Using a dexterous hand offers the benefits of multiple fingers with concurrent actions, which can be helpful for tactile exploration and pose estimation. However, tactile exploration with dexterous hands is challenging. The high-DOF configuration requires precise control to maintain effective and continuous contact with the object, ensuring that sufficient data is gathered throughout the process. Achieving this while maintaining sample efficiency--gathering valuable data without excessive interactions--is difficult.
Due to these complexities, many previous works rely on template-based methods \cite{Suresh22icra} to simplify the problem. Without object templates, accurate pose estimation requires more geometric-informed and sample-efficient policies to collect data.

Inspired by the human ability to reconstruct object shapes through touch and interactions alone, we propose a reinforcement learning (RL) based framework that enables robot hands to actively explore an object and refine its pose estimate with each new contact. Our approach uses the series of contact positions to iteratively build a 3D representation of the object. 
To enhance the effectiveness of this long-horizon exploration, we design the reward function to guide the robot toward contact points that maximize coverage and minimize pose uncertainty. This encourages the robot to develop a touch-based exploration policy that converges to accurate pose estimates over time. Additionally, we incorporate demonstration trajectories to initialize the system with a better heuristic. This accelerates training and helps the robot achieve efficient exploration faster.

In summary, our contributions include:
\begin{itemize}
\item A bimanual exploration pipeline that endows tactile-only object pose estimation with simple FSR sensors, enabling adaptability to objects of different sizes.
\item Designing novel state representation and reward function that promotes pose estimation while exploring the object within a limited number of actions.
\end{itemize}
We show that we can achieve 87\% accuracy on the ADD-S pose estimation metric \cite{Xiang2017PoseCNNAC} on unseen objects only through tactile exploration in 100 steps.

\begin{figure}[t]
    \centering
    \captionsetup{type=figure}
    \includegraphics[width=\linewidth]{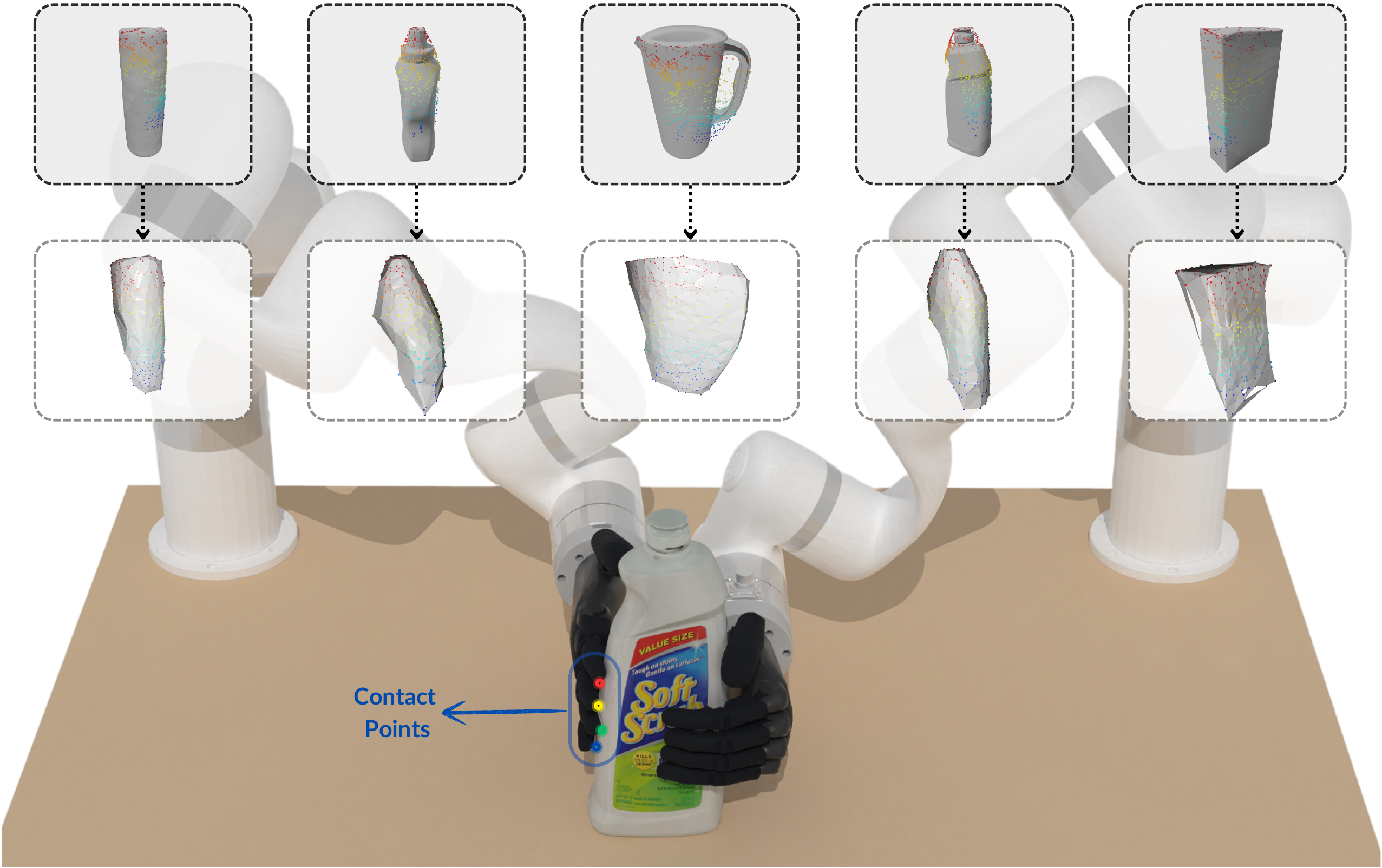}
    \caption{We present a tactile-only pose estimation framework using simple Force Sensing Resistors. We demonstrate that bimanual robot arms with dexterous hands can effectively explore \textbf{\small{[top row]}} and partially reconstruct \textbf{\small{[bottom row]}} the object's surface to estimate its pose.}
    \vspace{-0.7cm}
    \label{fig:teaser_fig}
\end{figure}

\section{Related Work}
\label{sec:related_work}
\textbf{Pose estimation}
Accurate object pose estimation is crucial for robot manipulation tasks~\cite{zeng2017multi, lin2024learning, wang2024lessons}. Vision-based approaches have proven successful in pose estimation~\cite{wen2024foundationpose, li2023nerf}, yet these methods are sensitive to environmental factors such as lighting conditions, shadows, occlusions, and reflectivity or transparency~\cite{bian2023transtouch, bauer2024challenges, zhang2022transnet}.
In contrast, tactile sensors provide contact-based feedback by measuring forces or deformations~\cite{yuan2017gelsight, lambeta2020digit, donlon2018gelslim, shahidzadeh2024feelanyforceestimatingcontactforce}, making them especially valuable in scenarios where vision systems face limitations~\cite{yin2023rotating, suresh2023neural, suresh2022shapemap}. Furthermore, compact tactile sensors can be embedded into end-effectors, offering a more integrated and self-contained sensing system compared with external cameras. ‌Recent works~\cite{10160359, suresh2023neural, Bauz2020TactileOP} have investigated pose estimation with vision-based tactile sensors, demonstrating promising results for object recognition and manipulation. however, these approaches typically rely on passive contact or pre-defined interactions, lacking active tactile exploration. This limitation restricts the range of objects that can be effectively explored, particularly those with complex geometries or varying sizes, and reduces the ability to gather rich, discriminative tactile information necessary for accurate pose estimation.  We overcome this challenge by addressing both \textit{active exploration} and \textit{pose estimation} in a bimanual setting, enabling the exploration of objects with varying geometric properties.

\textbf{Active perception} 
Traditional active sensing with vision-based methods primarily involves optimizing camera pose for better viewpoints~\cite{connolly1985determination, chen2011active}, which give better visual coverage and depth information, providing geometric information for tasks such as object recognition, mapping, and navigation~\cite{yu2024maniposecomprehensivebenchmarkposeaware}.
In recent years, tactile-based active sensing has gained a broader impact. By planning where to make contact, tactile sensors provide detailed local surface information about the object. Such coordinated sensing approaches enable robots to gather information effectively optimized for a given task~\cite{smith2021active, kim2022active}. However, due to the sparse tactile sensor input, high accuracy regarding geometric information may require object templates~\cite{xu2022tandem}. 

\textbf{Sensorimotor based exploration}
By leveraging sensorimotor coordination, robots can gather valuable sensory data more efficiently. Sensorimotor-based exploration has been extensively applied in active SLAM in object manipulation~\cite{dragiev2013uncertainty, pinto2016curious}. Earlier approaches to coordinating visual and tactile sensing in robotic hands predominantly relied on methods including information gain optimization~\cite{delmerico2018comparison}, uncertainty~\cite{ bjorkman2013enhancing, dragiev2013uncertainty}, potential fields~\cite{bierbaum2008potential}, Gaussian processes~\cite{martinez2017active, driess2017active, gandler2020object}.

Recent advancements in deep learning have introduced more sample-efficient approaches~\cite{schoettler2020deep}, enabling optimization for specific tasks such as object reconstruction~\cite{shahidzadeh2024actexplore, comi2024touchsdf, bimbo2022force}, recognition and classification~\cite{xu2022tandem}, shape completion~\cite{rustler2022active, rustler2023efficient, dutta2024vitract}, and pose estimation~\cite{liu2023enhancing}. However, exploration tasks are inherently long-horizon problems, and leveraging temporal multi-fingered data over time is essential. To highlight the importance of pose estimation feedback during training, we compare our method against a multi-fingered extension of a recent tactile exploration approach~\cite{shahidzadeh2024actexplore} in Sec.~\ref{sec:experiments}. Furthermore, most prior methods assume the object is fixed in the global frame by external forces; we address this limitation using a bimanual setup, where one hand holds the object during exploration.

\section{Problem Formulation}
\label{sec:problem_formulation}
We focus on the challenging task of bimanual tactile exploration, where one hand holds the object to provide a stable grasp for the other hand to explore. This setup requires each hand to be aware of the other hand's position to effectively operate within its own workspace without entering the other hand's workspace. Additionally, the exploring hand must make efficient, long-term decisions to extract pose-related information in as few steps as possible by leveraging its exploration history. Achieving this involves:
\begin{enumerate}
    \item \textbf{Initialization:} Both hands move toward the center of the workspace until they contact the object.
    \item \textbf{Holding Adjustment:} The holding hand finds a stable grasp pose on the object, while the exploring hand provides support from the opposite side.
    \item \textbf{Exploration:} Once stabilized, the exploring hand begins collecting contact data from its FSR sensors distributed around the fingertips.
\end{enumerate}
To tackle this problem, we have simulated the Ability Hands (discussed in Sec. \ref{sec:experiments}
) which facilitates the collection of contact points, $\mathbf{P}$, from designated FSR sensors along with estimated normals, $\mathbf{N}$, computed based on the hand's kinematics, enabling fast, safe interaction with objects to accommodate training holding and exploration policies. The collected point cloud data, expressed in the robot's base frame, is denoted as follows:
\begin{align*}
\mathbf{P} &= \{ \mathbf{p}^i = (x_i, y_i, z_i) \mid i \in \{1, 2, \dots, n\} \} \\
\mathbf{N} &= \{ \mathbf{n}^i = (n_{i_x}, n_{i_y}, n_{i_z}) \mid i \in \{1, 2, \dots, n\} \}
\end{align*}
\begin{figure}[t]
    \vspace{0.3cm}
    \centering
    \includegraphics[width=0.8\linewidth]{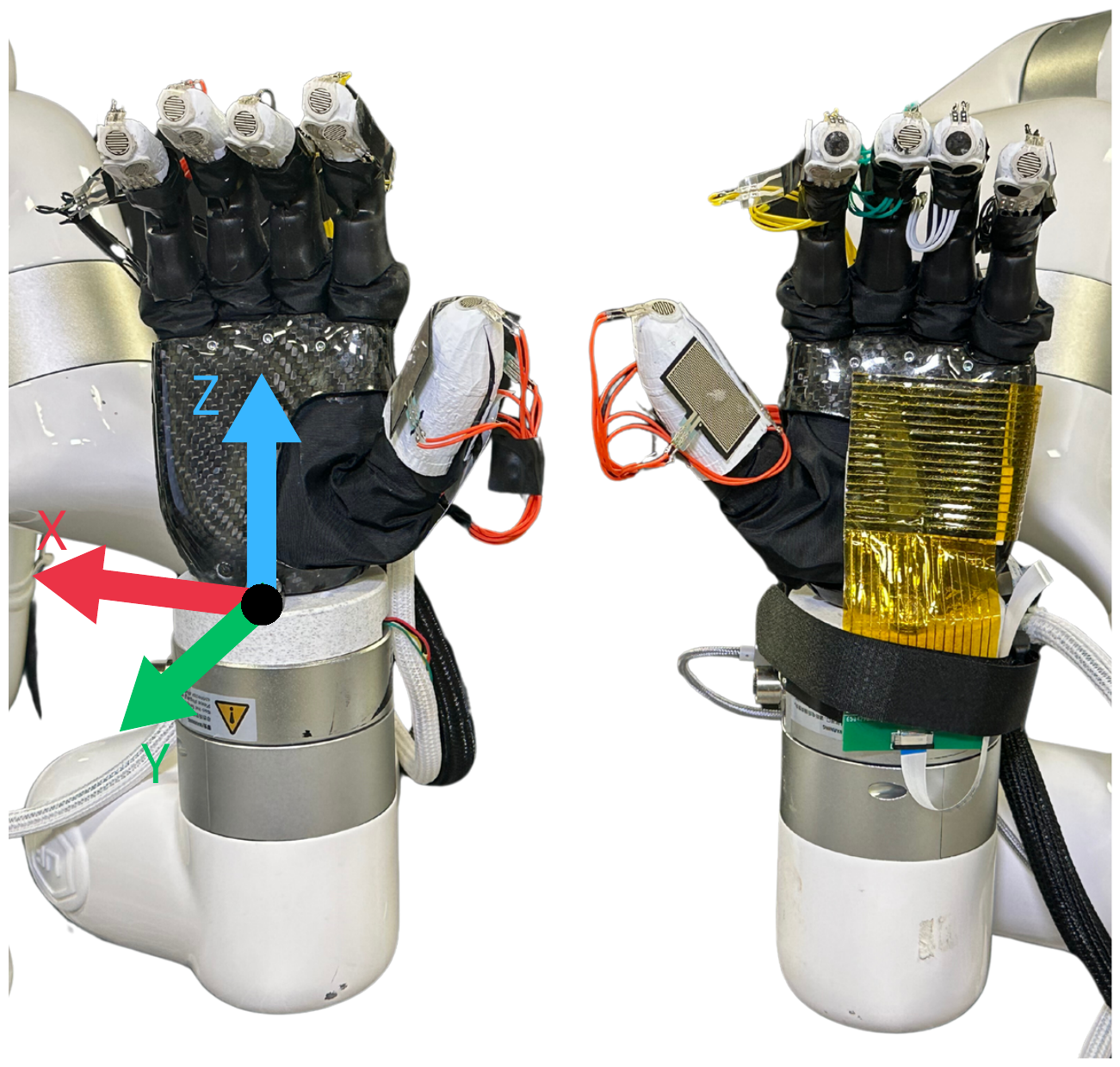}
    \caption{The Ability Hands prototype equipped with FSR sensors on the fingers to improve touch sensitivity. A palm tactile sensor~\cite{huang20243dvitac} is also mounted on the holding hand (left). Local action axes are depicted on the right hand.
    }
    \label{fig:fsr}
    \vspace{-20pt}
\end{figure}

\vspace{-15pt}
\section{Methodology}
\label{sec:methodology}
We will discuss different components of the algorithm, including observations (Sec. \ref{sec:state_rep}) and reward function (Sec. \ref{sec:reward}) as well as pose estimation procedure (Sec. \ref{sec:pose_est}).

\begin{figure*}
\centering
\vspace{0.1cm}
\includegraphics[width=\linewidth]{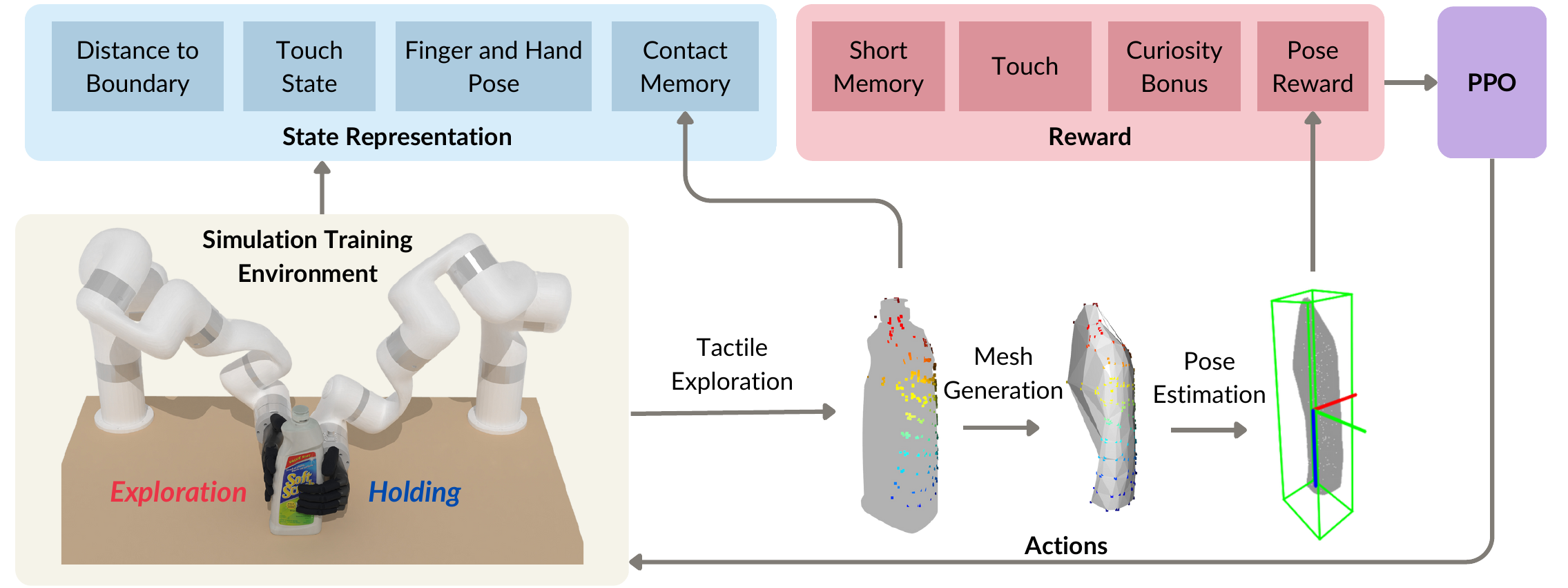}
\caption{The training framework of our approach. In the simulation, we have one hand exploring the object and obtaining binary tactile contact information. The relevant information is encoded in the state representation and rewards. During the process, we generate the mesh based on the point clouds. We then evaluate the pose and also include it in the reward function. The policy is trained with PPO on a small set of primitive objects.}
\vspace{-0.6cm}
\label{fig:framework}
\end{figure*}

\subsection{Pose Estimation}
\label{sec:pose_est}
Herein, we define our pose estimation pipeline using the contact point cloud $\mathbf{P}$ collected by the exploring agent. Inspired by recent advances in rigid-body pose estimation \cite{wen2024foundationpose, labbe2022megapose} using RGB-D images, we outline our approach to depth image rendering as follows. First, we reconstruct a surface from the point cloud \( \mathbf{P} \) using the ball-pivoting algorithm \cite{817351}, which has demonstrated promising results with point clouds containing missing points \cite{scutsurf_huang}—a common occurrence when collecting point clouds gathered with sparse sensor exploration. Then, we sample a new point cloud $\{\mathbf{P}^s, \mathbf{N}^s\}$ uniformly from this surface to obtain a dense point cloud for higher-resolution rendering. To render a depth image and ensure adequate visibility of points in $\mathbf{P}^s$, we generate $N_v$ viewpoints around the explorer hand's starting pose while facing the center of the workspace. Given each viewpoint pose $\{V_i\}_{i=0}^{N_v}$, we transform the point cloud $\{\mathbf{P}^s, \mathbf{N}^s\}$ into viewpoint frame $V_i$ which will be denoted as $\{\mathbf{P}^s_{V_i}, \mathbf{N}^s_{V_i}\}$. Visible points $\mathbf{P'}^s_{V_i}$ can be identified where
\begin{align*}
    \mathbf{P'}^s_{V_i} = \{ \mathbf{P}^s_{V_i}\mid \mathbf{P}^s_{V_i} \cdot \mathbf{N}^s_{V_i} < 0 \}
\end{align*}
A simple camera projection is performed on $\mathbf{P'}^s_{V_i} \in \mathbb{R}^{N \times 3}$ using the camera matrix $\mathbf{K}$, which calculates depth as $Z_{V_i} U_i = \mathbf{K}  {\mathbf{P'}^{s}_{V_i}}^T$,
where $U_i$ represents 2D pixel coordinates in homogeneous form and $Z_{V_i}$ contains the depth of the corresponding projected points along the viewpoint $V_i$. We then use FoundationPose \cite{wen2024foundationpose} to perform pose refinement and selection under model-based assumptions to determine object pose from the rendered depth images. Note that a mask of the object is not required, as we only render the depth image of the contact point clouds of the target object.

\subsection{Action Space}
\label{sec:action_space}
To enable agents to navigate a 3D workspace that includes all valid 6D poses, we define three axes in the wrist frame of the hand, allowing for movement along or opposite each axis, as well as rotation around each axis. Our action space is defined relative to the hand frame (Fig. \ref{fig:fsr}) rather than a fixed global frame since sensor readings are also hand-relative. For example, when using the right hand, if only the thumb is in contact with an object while the other fingers remain untouched, it suggests that the object is positioned to the left of the hand. In contrast, if actions were defined in a global frame, a \emph{move left} action would depend on the current wrist pose. Additionally, to control finger joints, we set specific bending thresholds; fingers will continue bending after each action until one of the sensors on the finger activates (minimizing the interaction with the object) or it reaches its bending threshold. The bent finger joint angles will then be used as part of observation as explained in Sec. \ref{sec:state_rep}. After recording each finger's joint angle and touch state, the fingers reopen to avoid sliding/shear force on the object while executing the next action. This idea is inspired by the fact that with each finger having one revolute joint and one passive joint, only a single grasp pose is possible per hand pose. Therefore, to control the hand's 6D movement, we have 12 actions: 6 translation actions and 6 rotation actions, where each action either increases or decreases translation or rotation along a specific axis defined on the hand's wrist.

\subsection{State Representation}
\label{sec:state_rep}
For an explorative policy, the state representation should capture information that enables the agent to make informed decisions about its next action, which explores the unvisited area. To this end, our state representation includes the following observations:

\textbf{Fingers Configuration} consists of finger joint angles, providing insight into the fingers’ current configuration and indirectly capturing geometric information about a specific part of the object in contact. As the fingers close around an object, each joint angle bends until it touches it. Thus, by observing patterns in the joint angles alone, one can infer local geometric features, such as curvature, as the hand adapts its grip to match the contact shape. We denote the finger joint angles $\mathcal{F}_t \in \mathbb{R}^5$ at time step $t$ as below
\begin{align*}
    \mathcal{F}_t = (q_1, q_2, q_3, q_4, q_5)
\end{align*}
where $q_1, q_2, q_3, q_4, q_5$ are the joint angles of thumb, index, middle, ring, and little fingers respectively at time step $t$ when the fingers are either touching the object or fully bent to a pre-defined threshold. 

\textbf{Hand Rotation} captures the wrist's orientation relative to its initial pose, allowing controlled movements that minimize unintended rotations that collide with the workspace boundaries. This helps the hand explore safely and reduces collision risks. We denote this orientation as $\mathcal{R}_t = ( \alpha, \beta, \gamma )$, 
where $\mathcal{R}_t \in \mathbb{SO}^3$ and $\alpha, \beta,$ and $\gamma$ are rotations at time $t$ about the x, y, and z axes w.r.t the initial wrist frame, respectively.

\textbf{Touch State} indicates which part of each finger-- fingertip, bottom, left or right side--is in contact with an object. This precise information allows the robot to assess not only if contact is made but also give insight into the alignment of the hand with the object surface in combination with other observations. We define it as a one-hot vector for each finger. For each finger \( f_i \) (where \( i = 1, \dots, 5 \)), we have vector \( \mathbf{t}_i \in \mathbb{B}^4\) representing the touch states:
\[
\mathbf{t}_i = \begin{bmatrix} t_{\text{tip}} & t_{\text{bottom}} & t_{\text{left}} & t_{\text{right}} \end{bmatrix}^T
\]
where each element \( t_{\text{.}} \in \{0, 1\} \) indicating whether a side is in contact or not. Thus, the entire hand’s touching state \( \mathcal{T}_t = [\mathbf{t}_1^T, \cdots, \mathbf{t}_5^T]^T \in \mathbb{B}^{20} \)  is then represented by concatenating the vectors for all five fingers at time $t$.

\textbf{Workspace Boundary} To restrict each agent’s workspace within predefined boundaries for safe exploration and prevent interference with other agents, we define the workspace using a set of boundary planes:
\label{sec:workspace}
\[
\mathbf{B}_i: a_i x + b_i y + c_i z + h_i = 0
\]
where \(\mathbf{n}_i = (a_i, b _i, c_i) \) is the normal vector of the plane and $h_i$ is the distance from the origin to the plane along \(\mathbf{n}_i\). At time $t$, given the agent’s wrist position $\mathbf{w}_t = (x_t, y_t, z_t)$ and a candidate action direction $\mathbf{d}_t^a$, the intersection of this direction with each opposing boundary plane ($\mathbf{n}_i \cdot \mathbf{d}_t^a < 0$) is computed as:
\[
\mathbf{p}_{ti}^a = \mathbf{w}_t + k \mathbf{d}_t^a, \quad k = -\frac{\mathbf{n}_i \cdot \mathbf{w}_t + h_i}{\mathbf{n}_i \cdot \mathbf{d}_t^a}.
\]

Then, the maximum allowable movements $k$ along action $a$ is then determined by the closest intersection:
\[
D^a_t = \min_i \| \mathbf{p}_{ti}^a - \mathbf{w}_t \| = \min_i |k| \, \|\mathbf{d}_t^a\|.
\]

This contributes to a boundary observation of $\mathcal{B}_t = \{D^a_t\}_{a\in [0,5]} \in \mathbb{R}^6$ providing the agent with a concise representation of workspace along each action direction.

\textbf{Local Contact Memory}
This observation provides the agent with the history of all the contact points, forming a point cloud $\mathbf{P}^i_t$, up to time $t$. This information can guide the agent to explore more effectively, minimizing revisits to previously explored areas. To achieve this, we first select the points near the current wrist pose $w_t$ and apply voxel grid filtering to approximate all points in every voxel grid with their voxel centroid to represent the surface more uniformly. 
\begin{gather*}
    \mathbf{P}'_t = \left\{\mathbf{P}_t^i \mid T > \|\mathbf{P}_t^i - w_t \| \right\} \\
\hat{\mathbf{P}}_t = \mathrm{VoxelGrid}(\mathbf{P}'_t)
\end{gather*}
We will then project them along every action axis $d_t^a$ to calculate $M_t^a$, which indicates the extent to which the area has been explored in the direction of action 
$a$.
\begin{gather*}
    M^a_t = \sum_k (\hat{\mathbf{P}}^k_t \cdot \mathbf{d}^a_t) \, \mathbb{I}(\hat{\mathbf{P}}^k_t \cdot \mathbf{d}^a_t > 0)
\end{gather*}
This observation can play a key role in choosing the near-optimal action when several reasonable actions allow the agent to explore the surface rather than free space; however, using $M_t$, the agent can make smarter decisions, leading to faster and more efficient exploration. Consequently, this observation is particularly useful for translational action agent moves left, right, up, or down.
\begin{align*}
    \mathcal{M}_t = (M^{\mathrm{LEFT}}_t, M^{\mathrm{RIGHT}}_t, M^
    {\mathrm{UP}}_t, M^{\mathrm{DOWN}}_t)
\end{align*}
resulting in local contact memory observation $\mathcal{M}_t \in \mathbb{R}^4$. Note that the lower the $M^a_t$, the less explored that area is for the agent, so the agent should prioritize actions that move toward the regions with smaller $M_t$. For further discussions on its effectiveness, please refer to Sec.~\ref{sec:experiments}.  
Our state at time $t$ is a concatenation of the above-mentioned observations 
 \begin{align*}
     S_t = \{ \mathcal{F}_t, \mathcal{R}_t, \mathcal{T}_t, \mathcal{B}_t, \mathcal{M}_t \}
 \end{align*}   

\subsection{Reward Design}
\label{sec:reward}
Our multi-objective reward function balances task-specific rewards for accurate pose estimation with intrinsic rewards that incentivize exploration and prolonged object interaction.

\textbf{Touch ($R^f_T$)}
 This intrinsic reward counts the number of fingers in contact at each time step. The idea is that more contact points indicate \emph{better hand alignment} with the object’s surface for a given pose, allowing the agent to gather more geometric information from the object. This reward is assigned for each finger $f$ using the corresponding touch state observation $R^f_T = \bigvee_s t_s$ where $s$ is the index of each sensor on the finger $f$.

\textbf{Curiosity Bonus ($R^f_B$)}
Drawing inspiration from the curiosity-driven reward function in \cite{shahidzadeh2024actexplore}, we defined an exploration bonus \textbf{per finger} to encourage exploration throughout the trajectory. This intrinsic reward incentivizes the agent to discover new positions by keeping track of all the positions each finger has visited $N_t(f, \mathrm{T}_t)$. Locations visited fewer times should receive a higher reward, promoting exploration of unvisited areas. Thus, for curiosity-driven reward, we'll have: 
\begin{align*}
    R^f_B = \frac{1}{N_t(f,\mathrm{T}_t)}
\end{align*}
where $f \in [0, 1, \ldots, 4]\}$ is the finger index and $\mathrm{T}_t$ is the position the finger visits at time $t$. Defining this reward on a per-finger basis rather than per-hand position allows each finger to reach the same positions from various orientations, with the possibility that other fingers may touch new positions based on the object’s shape. 

This reward is positively correlated with the IoU of the contact point cloud $\mathbf{P}$ and the object. However, since there are various exploration patterns that can lead to increased IoU, an additional extrinsic reward $R_P$ is introduced to guide the exploration policy toward more accurate pose estimation.

\textbf{Short Memory ($R_M$)}
To effectively utilize the history $\mathcal{H}$ of observations and actions, we used a penalty term based on the last $k$ poses and actions taken by the agent. Specifically, the agent receives a negative reward $R_M$ if it repeats the same action in the same pose within those $k$ steps. This approach prevents the agent from falling into short-horizon loops, where it learns a sequence of actions to receive $R_B$ without engaging in long-horizon exploration. Our experiments indicate that this reward encourages the agent to adopt a faster exploration strategy.

\textbf{Pose Estimation Feedback ($R_P$)}
 To encourage trajectories that gather more pose-related information from objects, we use pose estimation accuracy as a feedback signal for the exploration policy. This approach prioritizes not only exploration of the object but also efficient pose estimation in fewer actions. To this end, we use the AUC of ADD-S~\cite{Xiang2017PoseCNNAC} after performing mesh reconstruction and depth image rendering, as explained in Sec.~\ref{sec:pose_est}, from partial reconstruction point cloud $\mathbf{P}_t$ at time $t$. Therefore, for $R_P$ we'll have
\begin{align*}
     R_P = \text{AUC}_{\text{ADD-S}}(\mathbf{P}_t, \mathcal{P}_{GT}) \enspace \mathbb{I}(t \text{ mod } C)
\end{align*}
where $\mathcal{P}_{GT}$ represents the ground truth pose from the simulator, and $\text{AUC}_{\text{ADD-S}}$ estimates the pose estimation accuracy.  Given the computational cost (7 to 15 seconds) involved in calculating this reward, as well as the minimal changes observed at each step, we calculate the reward every $C$ steps (we set $C=50$ in our experiments).  Additionally, we fine-tune its regularizer to ensure consistent effectiveness compared to other rewards.

Eventually, our multi-objective reward function will be formulated for each finger as:

\begin{equation*}
R_f= 
    \begin{cases}
        R^f_T R_M, & \text{ if}  (\mathcal{P}_t, a_t) \in \mathcal{H}         \\ %
                R^f_T(\alpha + \beta R^f_B), &   \text{ow.} 
    \end{cases}    
\end{equation*}
And the total reward for $(s_t, a_t)$ pair is as follows:
\begin{align*}
    R_t &= \frac{\sum_{f\in [0,4]} R_f}{5} + \gamma R_p
\end{align*}

\vspace{-0.3cm}
\section{Experiments}
\label{sec:experiments}

\textbf{Experiment Setup}
We evaluate our framework in simulation using the 6-DoF Ability Hands modeled in SAPIEN \cite{Xiang_2020_SAPIEN}. Each fingertip is divided into five sensing regions (top, bottom, left, right, and tip) which will be activated based on the specified boundaries. In the bimanual setup, both hands move toward a randomly posed object oriented toward the palms, ensuring contact and avoiding free-space exploration. Once the exploring hand makes initial contact, it remains stationary to provide support, while the holding hand adjusts to achieve a stable grasp. After the grasp is established, workspace boundaries for the exploring hand are defined, explorer hand begins active exploration. Our simulation-based evaluation effectively demonstrates the approach and enables comparison of different policy performances in a controlled setting. Therefore, real-world experiments were not the primary focus of this work.

 \begin{algorithm}[h!]
\footnotesize

\caption{\textsc{Explorer Training}}\label{alg:overall}
\DontPrintSemicolon
\For{episode = $1,2, \dots$}{
    $\mathcal{H} \gets$ history of size $k$\;
    $N(f,T) \gets 0$ for all fingers $f$ and $T \in$ workspace\;
    \For{$t = 0, 1, 2, \dots, T-1$}{
        $s_t \gets \{ \mathcal{F}_t, \mathcal{R}_t, \mathcal{T}_t, \mathcal{B}_t, \mathcal{M}_t \}$\;
        $a_t \gets \arg\max_{a'} \pi_{\theta}(a'|s_t)$\;
        $\mathcal{P}_{t+1} \gets$ explorer.step($a_t$)\;
        explorer.bendFingers()\;
        $t' \gets t+1$\;
        $\mathbf{P}_{t'} \gets$ updateContacts()\;
        $\mathcal{F}_{t'}, \mathcal{R}_{t'}, \mathcal{T}_{t'}, \mathcal{B}_{t'}, \mathcal{M}_{t'} \gets$ updateStates()\;
        explorer.openFingers()\;
        
        \For{$f = 0, 1, 2, \dots, 4$}{
            $t_f \gets$ finger pose $f$\;
            $R^f_T \gets \bigvee_s t_s$\;
            $R_f \gets 0$\;
            
            \uIf{$R^f_T = 0$}{
                \textbf{continue}\;
            }
            \uElseIf{$(\mathcal{P}_{t} , a_t) \in \mathcal{H}$}{
                $R_f \gets R_{M}$ \tcp*{\small{short memory}}
                \textbf{continue}\;
            }
            \Else{
                $R^f_C \gets \frac{1}{N_t(f, t_f)}$\;
                $N_t(f, t_f) \gets N_t(f, t_f) + 1$\;
                $R_f \gets \alpha R^f_T + \beta R^f_C$\;
            }
        }
        
        $R_p \gets \text{AUC}_{\text{ADD-S}}(\mathbf{P_{t+1}}, \mathcal{P_{GT}})$\;
        $R_t \gets \frac{\sum_{f\in [0,4]} R_f}{5} + \gamma R_p$\;
        $\mathcal{H}_{t+1} \gets \mathcal{H}_{t} \cup (\mathcal{P}_t, a_t)$\;
        
        \If{\text{terminationCheck}()}{
            \textbf{continue}\;
        }
    }
}
\end{algorithm}

\textbf{Training and Test Objects} Training objects
selected for their varying sizes and surface types (flat, corner, edge, and curved), are based on the concept that all object surfaces can be represented as combinations of primitive shapes (cuboid, cylinder, sphere). Test objects are drawn from the YCB dataset, including everyday items with convex, concave, long, and wide shapes. We have provided holding and exploration trajectories in the accompaniying videos.

\textbf{Explorer Policy}
Bimanual dexterous exploration involves complex, long-horizon control, requiring a well-designed state representation and reward function. We model the problem as a POMDP and train our policy using PPO with discrete actions. We began each episode by initiating contact with the object from random poses to start on-policy PPO training. To mitigate unusual behaviors during exploration, we reset the exploration episodes if:
\begin{enumerate}
    \item the explorer leaves the workspace or exceeds orientation thresholds (Boundary Check)
    \item the agent loses contact for more than 5 steps
    \item the maximum horizon of 400 steps is reached.
\end{enumerate}

\begin{table*}[ht!]
\centering
\vspace{0.2cm}
\caption{Quantitative evaluation of reward designs including (\textbf T)ouch with short (\textbf M)emory, Curiosity (\textbf{B})onus, and (\textbf P)ose estimation.
TMBP achieves the most accurate pose estimation, highlighting the importance of pose estimation feedback in guiding exploration behavior.}
\begin{small}
\setlength\tabcolsep{7pt}
\setlength{\extrarowheight}{1.5pt}
\scalebox{0.8}{

\begin{tabular}{c*{14}{>{\columncolor{gray!20}}c>{\columncolor{white}}c}}
\Xhline{1pt}
\multirow{2}{*}{} 
 Objects & \multicolumn{2}{|c|}{mustard bottle} & \multicolumn{2}{|c|}{chips can} & \multicolumn{2}{|c|}{pitcher vase} & \multicolumn{2}{|c|}{bleach cleanser} & \multicolumn{2}{|c|}{mug} & \multicolumn{2}{|c|}{sugar box} & \multicolumn{2}{|c}{Average}\\
  & \multicolumn{1}{|c}{IoU $\uparrow$} & \multicolumn{1}{c|}{ADD-S $\uparrow$} 
 & \multicolumn{1}{|c}{IoU} & \multicolumn{1}{c|}{ADD-S} 
 & \multicolumn{1}{|c}{IoU} & \multicolumn{1}{c|}{ADD-S} 
 & \multicolumn{1}{|c}{IoU} & \multicolumn{1}{c|}{ADD-S} 
 & \multicolumn{1}{|c}{IoU} & \multicolumn{1}{c|}{ADD-S} 
 & \multicolumn{1}{|c}{IoU} & \multicolumn{1}{c|}{ADD-S} 
 & \multicolumn{1}{|c}{IoU} & \multicolumn{1}{c}{ADD-S}\\
 \Xhline{0.8pt}
Grid Search &0.420 & 0.591 &0.439 & 
   0.835 &0.350 & 0.817 &0.462 & 0.807 &0.457  &0.915  &0.394  &0.665 &0.420 &0.772 \\

TMBP \textbf{(ours)} &\textbf{0.643}  &0.845  &0.589  &0.885  &\textbf{0.508}  &0.683  &\textbf{0.726}  &0.943  &0.452  &0.915  &0.596 &\textbf{0.971}  &0.586 &\textbf{0.874} \\
TMB &0.567  &0.734  &\textbf{0.632}  &0.843  &0.499  &0.633  &0.697  &\textbf{0.958}  &\textbf{0.566}  &0.896  &\textbf{0.621}  &0.864  &\textbf{0.597} &0.821 \\
TB &0.506  &0.752  &0.528  &\textbf{0.934}  &0.366  &\textbf{0.710}  &0.555  &0.935  &0.508  &0.896  &0.391  &0.897  &0.476 &0.854 \\
TM &0.559  &\textbf{0.860}  &0.521  &0.477  &0.423  &0.645  &0.570  &0.942  &0.448  &\textbf{0.933}  &0.540  &0.939  &0.510 &0.799 \\
\Xhline{1pt}
\end{tabular}
}
\end{small}
\vspace{-4pt}
\label{tab:reward_ablation}
\vspace{-5pt}
\end{table*}

\textbf{Holding Policy}
Inspired by human bimanual exploration, where one hand stabilizes the object while the other explores, we defined distinct tasks for each hand. Holding the object is a short-term task, where one hand grasps the object to counteract forces from the exploration hand.
To train the holding task, we used imitation learning based on human demonstrations. The holding hand's observations included Finger Configuration ($\mathcal{F}_t$), Hand Orientation ($\mathcal{R}_t$), and Touch State ($\mathcal{T}_t$) as described in Sec.~\ref{sec:state_rep}, which were used to generate expert actions and build a dataset. To improve contact detection, additional FSR sensors were added to the palm, resulting in $\mathcal{T}_t \in \mathbb{B}^{24}$.

\textbf{Baselines}
To assess the contribution of different state and reward components, we report ablations in Tab. \ref{tab:reward_ablation} and Tab. \ref{tab:obs_ablation}, using abbreviations to denote each variant. For comparison with prior work that only incorporates the curiosity bonus~\cite{shahidzadeh2024actexplore} and primarily evaluates IoU, we extend this reward to a multi-finger setup as the TMB variant. In addition, we implement a grid search exploration baseline to represent non-learning strategies: the hand moves in a fixed two-dimensional grid pattern It first moves upwards to the top, then moves one step to the right, moves downward to the bottom, and continues moving right and upward, repeating this pattern while using rule-based methods to align the hand's orientation with the object’s surface.

\begin{figure}[b]
    \centering
    \vspace{-10pt}
    \includegraphics[width=\linewidth]{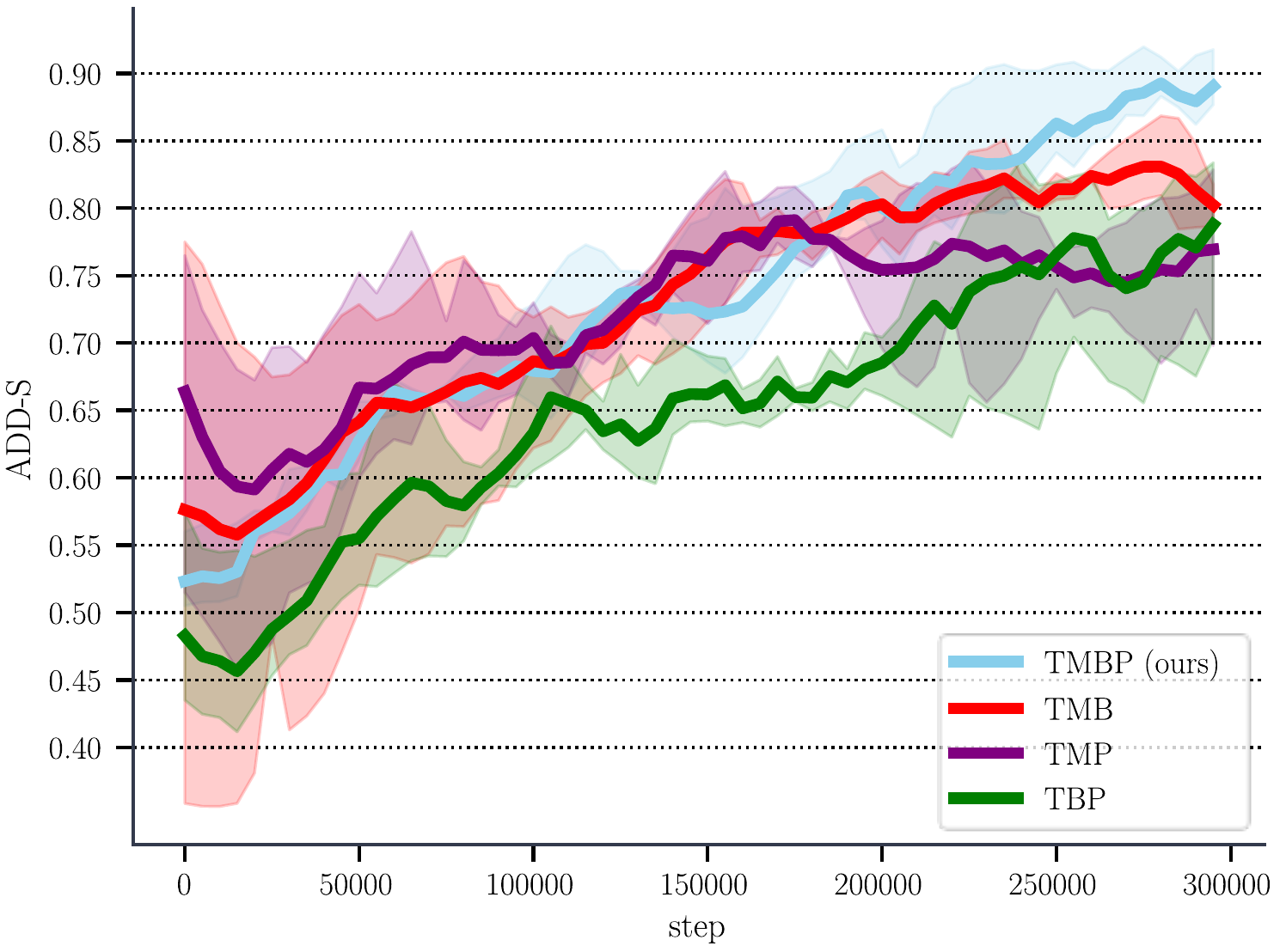}
    \caption{Pose estimation accuracy for different reward settings during training, averaged over three trainings with different seeds. All models use BFTRM as the state representation.}
    \label{fig:training_auc}
    \vspace{-10pt}
\end{figure}

\textbf{Training Findings}
 We assess the necessity of each component in our reward function by comparing the TMBP reward with other variants in Fig. \ref{fig:training_auc}:

\begin{enumerate}
    \item \textbf{TMBP vs. TMB:} Excluding pose estimation feedback makes the reward focus solely on exploration. While this leads to broader surface coverage, it fails to consistently improve ADD-S.
    \item \textbf{TMBP vs TMP:} Excluding the curiosity bonus leads to poor exploration behavior, with the agent repeatedly visiting similar regions. Consequently, pose estimation accuracy barely improves.
    \item \textbf{TMBP vs TBP:} Excluding the short-memory penalty makes the agent prone to revisiting the same poses and actions, producing inefficient, looped trajectories.
\end{enumerate}
Overall, we found that balancing exploration with an auxiliary objective like pose estimation requires all components, as each plays a distinct role. The short-memory penalty (M) and curiosity bonus (B) promote diverse and efficient exploration, while the pose estimation reward (P) directs this exploration toward task-relevant features. Removing any of these elements leads to degraded performance which highlights their complementary nature.  

\begin{figure}[h!]
        \centering
        \vspace{0.2cm}
    \includegraphics[width=0.95\linewidth]{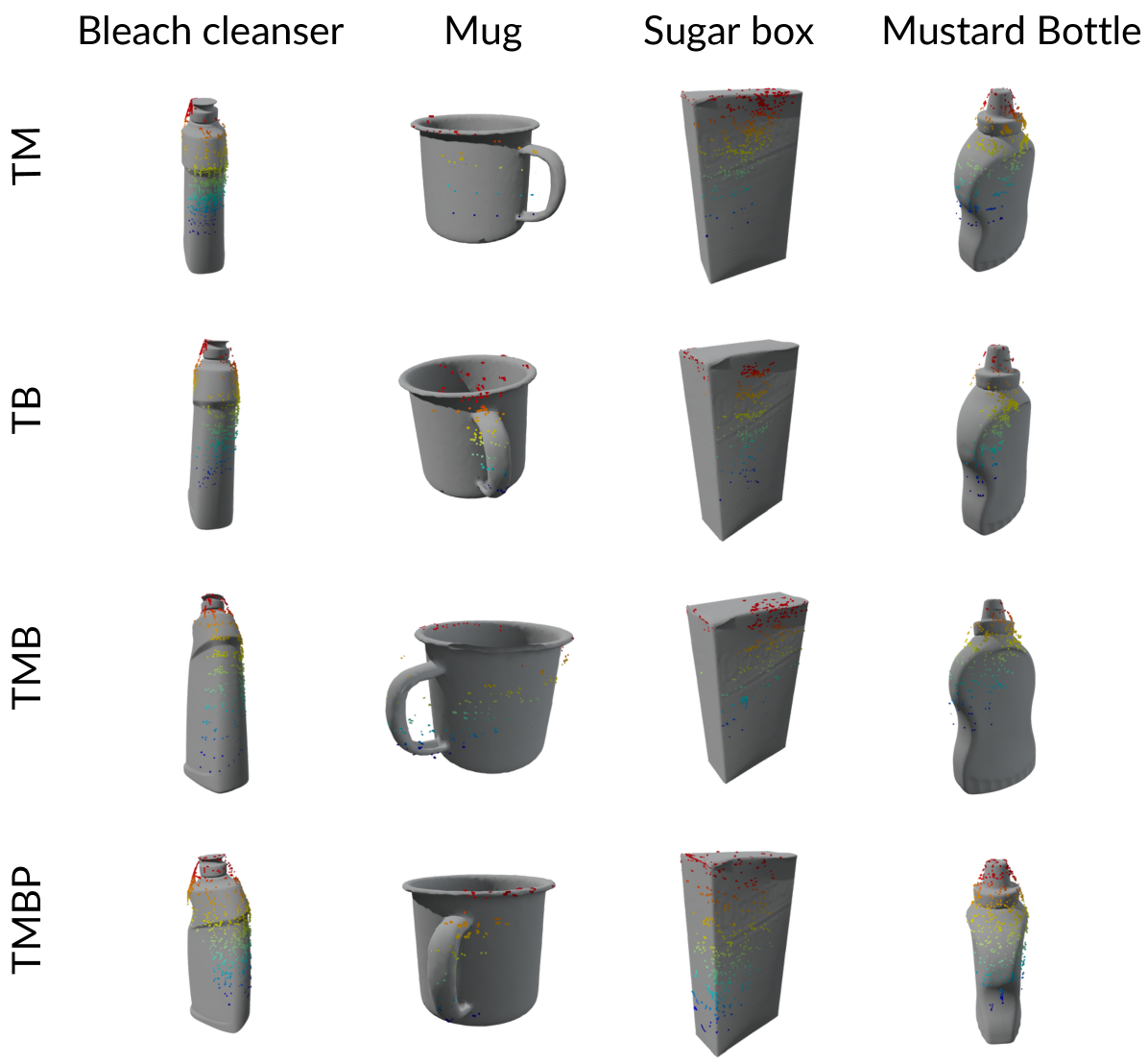}
        \caption{Qualitative results of exploration contact points, shown as colored points on unseen objects. We observe that our reward function effectively leverages pose estimation feedback to generate denser contact points on various target objects for accurate pose estimation. Each object is shown from the viewpoint with the highest number of contacts. TM, TB, and TMBP are different reward settings as explained in Sec. \ref{sec:experiments}. \small{(Best viewed by zooming in)}}
        \vspace{-0.6cm}
    \label{fig:qualitative}
\end{figure}

\textbf{Results and Discussions}
In pursuit of tactile-only object exploration for pose estimation, we run each test rollout up to 150 steps, depending on the object’s size, to ensure sufficient contact points were collected. Rollouts terminated either when the maximum step count was reached or a reset condition was triggered. For each model–object pair, we conducted four trials and report the average IoU and AUC of ADD-S in Tab. \ref{tab:reward_ablation} and Tab. \ref{tab:obs_ablation}. Tab. \ref{tab:reward_ablation} demonstrates the ablation study of different reward variants TMBP, TMB, TB, and TM which outperform the deterministic Grid Search baseline in terms of IoU and AUC of ADD-S metrics on test objects. Among these variants, TMBP achieves the highest average AUC of ADD-S which confirms our intuition that pose estimation feedback can promote exploring pose related features. In contrast, TMB attains the highest average IoU, indicating broader surface coverage and exploration completeness, consistent with the effect of the curiosity bonus \cite{shahidzadeh2024actexplore} and training findings.

\textbf{Reward Function} Comparing TMBP and TMB in Fig.~\ref{fig:qualitative} further illustrates this trade-off, TMBP explores objects by collecting contact points \emph{densely}, enabling a better local reconstruction and resulting in a higher AUC of ADD-S—except for the bleach cleanser, whose distinct features make it inherently easier for pose estimation overall. In contrast, TMB achieves a higher average IoU, reflecting more complete surface coverage. However, this demonstrates that \emph{IoU alone is not a reliable indicator of pose estimation performance, as higher IoU does not necessarily capture critical object features}. By incorporating pose-aware rewards, TMBP prioritizes exploration of discriminative features, thereby improving estimation accuracy even with lower overall IoU. For example, on the sugar box, TMBP focuses on the edges outlining the box shape, while on the mustard bottle it concentrates on the top curvature and lid. In contrast, TMB distributes exploration more sparsely across different sides to increase IoU, which is less suitable for the pose estimation algorithm described in Sec.~\ref{sec:pose_est}.

Comparing reward functions TB, TM, with TMBP, we observe that TB yields the lowest average IoU but the second-highest AUC of ADD-S, while TM performs poorly on both IoU and AUC of ADD-S due to its inability to account for visited areas, which increases the likelihood of revisits. Comparing TMB with TB further underscores that while the short memory mechanism promotes broader exploration, a higher IoU does not necessarily guarantee a higher ADD-S. Notably, TB also produces denser exploration contacts compared to TMB and TM (both of which apply the short-memory penalty), despite its lower IoU (Fig. \ref{fig:qualitative}).

\begin{table}[h]
\centering
\caption{IoU value of observation ablation on test objects. Note that BFTRM is a state representation which involves \textbf{B}oundary distance, \textbf{F}inger joints, \textbf{T}ouch State, Hand \textbf{R}otation, and local contact \textbf{M}emory}
\begin{small}
\setlength\tabcolsep{3pt}
\scalebox{0.64}{
\begin{tabular}{c*{7}{>{\columncolor{white}}c>{\columncolor{white}}c}}
\Xhline{1pt}
\multirow{1}{*}{Objects} 
 & \multicolumn{1}{c}{mustard bottle} & \multicolumn{1}{c}{chips can} & \multicolumn{1}{c}{pitcher base} & \multicolumn{1}{c}{bleach cleanser} & \multicolumn{1}{c}{mug} & \multicolumn{1}{c}{sugar box} &\multicolumn{1}{c}{Avg. IoU}
 \\ \Xhline{0.8pt}
Grid Search &0.420 &0.439 &0.350 &0.462 &0.457  &0.394 &0.420 \\ 
BFTRM \textbf{(ours)} &\textbf{0.567}  &\textbf{0.632}  &\textbf{0.499}  &\textbf{0.697}  &0.566  &\textbf{0.621}  & \textbf{0.597}\\
FTRM &0.545  &0.511  &0.329  &0.580  &0.617  &0.532 &0.519 \\
BFTR &0.549  &0.483  &0.363  &0.554  &\textbf{0.676}  &0.516 &0.524 \\ \cline{2-8} 
\Xhline{1pt}
\end{tabular}
}
\end{small}

\label{tab:obs_ablation}
\end{table}

\textbf{State Representation} Tab. \ref{tab:obs_ablation} shows exploration capability of our state representation vs three other variants, BFTRM, FTRM, and BFTR, evaluated using the IoU metric while using the TMB reward to better showcase the exploration objective. The results show that all variants outperform the deterministic Grid Search algorithm baseline on test objects, demonstrating the effectiveness of BFTRM richness. Among the variants, BFTRM achieves the highest average IoU, highlighting that it's more suitable for exploration. Comparing FTRM and BFTR, we observe that incorporating boundary distance is essential for avoiding collisions, while local contact memory enhances exploration efficiency within limited actions. Ultimately, our experiments demonstrate that each state and reward component is fundamental, and removing any of them adversely affects pose estimation objective. 

\vspace{-0.2cm}
\section{Conclusion}
\label{sec:conclusion}
In this work, we presented a reinforcement learning-based framework to actively touch objects and estimate the poses. We utilize the low-cost force-sensing resistors combined with bimanual dexterous robot hands. By using pose rewards and observation history, our method iteratively collects point clouds of the touch points and estimates the object poses. Our experiments show that by training only on five primitive shapes, we can effectively estimate a wide range of poses of daily objects. 
Looking forward, we hope to inspire broader applications of tactile exploration, such as sensing in-the-loop manipulation policies and exploiting tactile feedback from the holding hand for pose estimation.

\bibliographystyle{IEEEtran}
\bibliography{main}

\end{document}